\RequirePackage{fix-cm}
\documentclass[twocolumn]{svjour3}          
\smartqed  
\usepackage{graphicx}
\usepackage{amsmath}
\usepackage{array}
\usepackage{subfigure}
\usepackage{wrapfig}

\usepackage{hyperref}
\hypersetup{
	colorlinks=true,
	linkcolor=blue,
	filecolor=magenta,      
	urlcolor=cyan,
	citecolor=blue
}

\usepackage{multirow}
\usepackage{algorithm}
\usepackage[noend]{algpseudocode}

\usepackage[usenames,dvipsnames,svgnames,table]{xcolor}
\definecolor{junglegreen}{rgb}{0.16, 0.67, 0.53}
\definecolor{purple}{rgb}{0.53,0.16,0.88}
\definecolor{mypink1}{rgb}{0.858, 0.188, 0.478}

\usepackage{microtype}

\journalname{} 
\begin{document}\sloppy

\title{Robotic additive construction of bar structures:\\Unified sequence and motion planning
}


\author{Yijiang Huang$^1$
	\and
	Caelan R. Garrett$^2$
	\and
        Ian Ting$^3$
	\and
        Stefana Parascho$^3$
	\and
	Caitlin T. Mueller$^1$
}

\institute{Yijiang Huang \at
	\email{yijiangh@mit.edu}
	\and
	Caelan R. Garrett \at
	\email{caelan@csail.mit.edu}
	\and
	Ian Ting \at
	\email{ianting@princeton.edu} 
	\and
	Stefana Parascho \at
	\email{parascho@princeton.edu} 
	\and
	Caitlin T. Mueller \at
	\email{caitlinm@mit.edu} 
	\and
	$^1$Building Technology Program, Department of Architecture, Massachusetts Institute of Technology, Cambridge, MA, 02139 USA\\
	\\$^2$Computer Science and Artificial Intelligence Laboratory, Massachusetts Institute of Technology, Cambridge, MA, 02139 USA\\
        \\$^3$Princeton University, CREATE Laboratory, School of Architecture, Princeton, NJ, 08544 USA
}


\date{Received: date / Accepted: date}

\maketitle

\begin{abstract}

Additive robotic construction of building-scale discrete bar structures, such as trusses and space frames, is increasingly attractive due to the potential improvements in efficiency, safety, and design possibilities. However, programming complex robots, such as manipulators with seven degrees of freedom, to successfully complete construction tasks can be tedious, challenging, or impossible for a human to do manually. Namely, the structure must be constructed in a sequence that preserves structural properties, such as stiffness, at each step. At the same time, this sequence must allow for the robot to precisely manipulate elements within the in-progress structure while respecting geometric constraints that, for example, ensure the robot does not collide with what it has built. In this work, we present an automated and newly generalized planning approach for jointly finding a construction sequence and robot motion plan for additive construction that satisfies these requirements. Our approach can be applied in a variety of additive construction processes, and we demonstrate it specifically on spatial extrusion and discrete bar assembly in this paper.  We demonstrate the effectiveness of our approach on several simulated and real-world extrusion and assembly tasks, including a human-scale physical prototype, for which our algorithm is deployed for the first time to plan the assembly of a complicated double tangent bar system design.

\end{abstract}

\begin{figure}[h!]
  \centering
  \includegraphics[width=1\columnwidth]{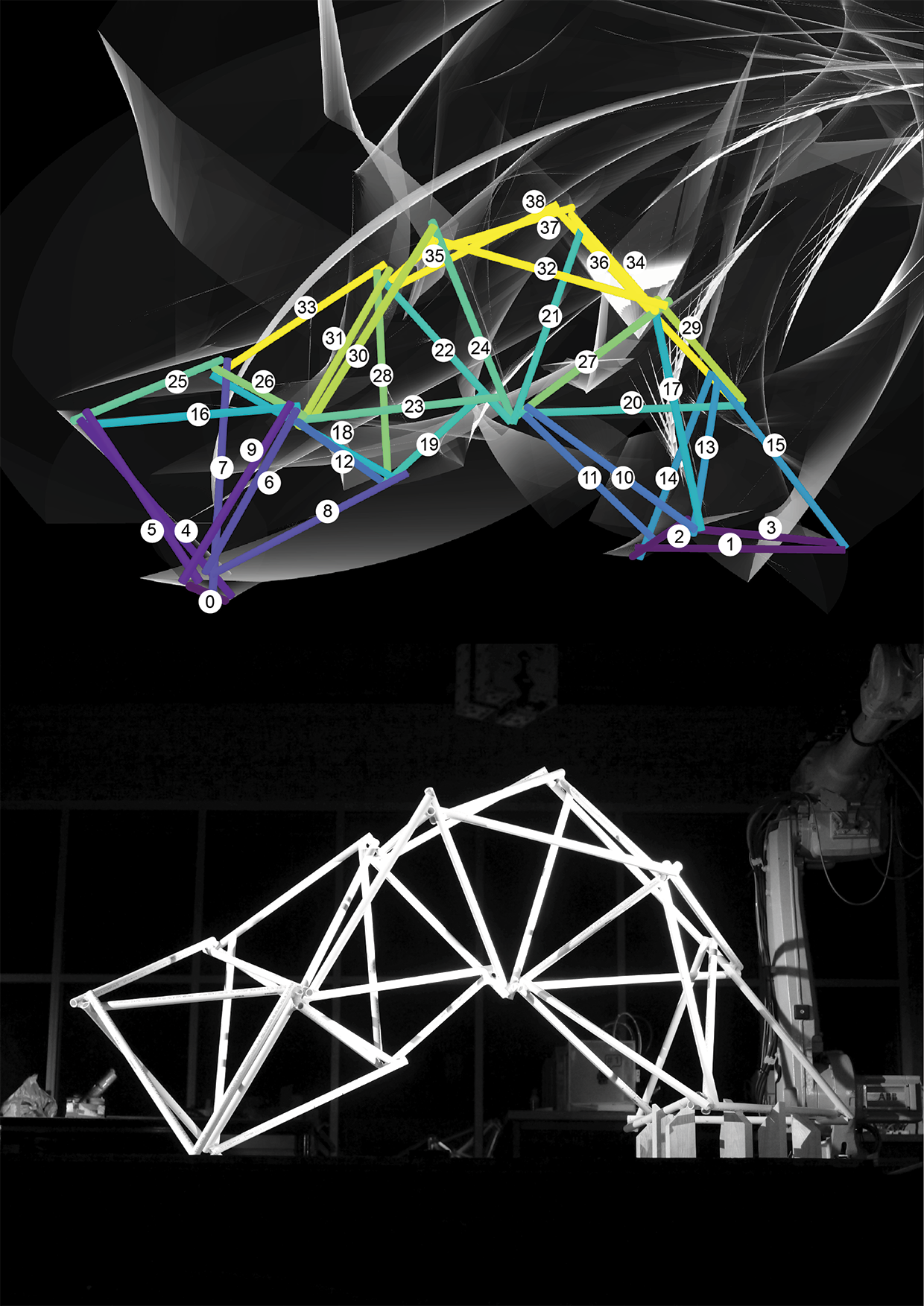}
  \caption{{\em Above}: construction sequence simulation. The sequence is marked by numbers and colors. The bars' sweeping envelop when grasped by the robot is also illustrated. {\em Below}: the real built example of a trussed arch structure, additively constructed by a robotic arm with the sequence and motion planned by our algorithm.}
 \end{figure}


\section{Introduction}\label{sec:intro}

Robotics-based fabrication promises to substantially improve construction in the built environment, offering benefits such as speed, quality, material efficiency, worker safety, and eventually cost reduction. In addition, robotic additive construction opens up new fabrication possibilities beyond automating manual processes, leading to an expansion of design possibilities and formal expressions. Recent years have started to see these decades-old promises finally come to fruition, primarily in the form of impressive prototypes, proofs-of-concept, and demonstrator projects at the "pavilion" scale, showcasing the potential of robotic fabrication and assembly in architecture. However, key challenges still remain to be tackled before the impact of these technologies are broadly felt on real construction sites. This paper addresses one such challenge: automated methods for planning and programming the robotic systems that will assemble building systems. In particular, this paper focuses on a broad class of building structures, bar structures, which are composed of discrete linear elements and used broadly and in great variety in the built environment. Trusses, frames, and more complex structural hierarchies involving beams and columns are all examples of bar structures, and can use a range of materials such as metals, timber, precast concrete, plastics, etc.

3D bar structures can have complex geometries and topologies, for reasons related to aesthetics, structural efficiency, material availability, or site constraints. Because of this complexity, robotic assembly, i.e. using industrial robotic arms to pick and place structural elements, is an attractive construction approach due to the dexterity and geometric range of industrial robots; there is the potential to precisely and efficiently construct highly tuned structures designed in response to a range of priorities and constraints. However, because of this design complexity, the challenge of robotically assembling such structures is high, and one can not typically fall back on rules of thumb or previous solutions~\cite{eversmann2017prefabtimber,sondergaard2016topology}. 
Previous work has demonstrated great potential for additive construction of bar structures using robots with many degrees-of-freedom (DOFs)~\cite{gramazio2014robotic,willmann2016robotic}, but also significant hurdles: planning the sequence and motion of these robots must simultaneously account for collision avoidance in dense bar networks and structural stiffness during construction. These computational challenges, coming from both the high DOFs of the machine and the large-scale structural behavior of the partially constructed structures, distinguish robotic additive construction from other layer-based additive manufacturing techniques. While transitioning between a volumetric digital design model and machine code for a 3-axis gantry machine is relatively straightforward, searching for a feasible construction sequence and trajectories for robotic arms is much less obvious, and thus requires a different planning strategy and new design methodologies.

Recent previous research has proposed scalable planning algorithms to address these constraints in the context of robotic spatial extrusion~\cite{garrett2020scalable}, finally empowering solutions to large robotic planning problems without the use of human intervention. However, more generalized assembly planning for additively assembled structures has not been addressed.

In this paper, we generalize previous work on extrusion planning~\cite{garrett2020scalable} to demonstrate how to solve a variety of  additive assembly planning problems using the same search strategy. In particular, we use the concepts of an {\it action template} and {\it plan skeleton}, which together describe a sequence of robot’s motion primitives in these two different applications. Formalizing additive construction methods such as extrusion and assembly planning through the lens of a plan skeleton allows us to use the same planning algorithm to solve a range of construction problems efficiently, improving automation of additive construction planning in general. We demonstrate the effectiveness of our approach through simulation and real-world construction examples.

\section{Related work}\label{sec:related_work}

Automating the construction of bespoke, irregular bar structures using programmable robots has been studied in the rapidly advancing field of architectural robotics in the past two decades, specifically due to a robot’s capacity for performing precise spatial movements. 
Spatial extrusion and robotic assembly  are the two main types of methods that use a robot to directly distribute (by extruding or positioning) individual, often standardized linear elements in positions designated by the structure’s design. 
Existing pilot studies have demonstrated the great potential of deploying this technology to the scale of a building~\cite{eversmann2017prefabtimber,hack2014mesh,helm2015iridescence,thoma2018robotic}, allowing both formal variations~\cite{soler2017generalized,yuan2016robotic} and structural efficiency~\cite{sondergaard2016topology,tam2018}. However, much of this early work adopts a trial-and-error method for planning, by manually conjecturing a construction sequence and end effector workspace poses for the robot. Software packages exist to support these methods by performing point-wise kinematics checks~\cite{braumann2011parametric,schwartz2012hal} or configuration-to-configuration motion planning~\cite{gandia2018towards,sucan2013moveit}. However, these tools currently require a sub-optimal manual planning process because they cannot reason about the construction sequence and trajectories simultaneously. This manual requirement limited early work to bar structures with repetitive topological patterns.

Several sequence and motion planning algorithms have been proposed to robotically extrude bar structures with arbitrary geometry and topologies. Early work in this direction addressed planning for a free-flying hotend end effector~\cite{huang2016framefab,wu2016printing,yu2016acadia}. 
By rigidly committing to a partial ordering on the construction sequence, these algorithms are incomplete. Gelber et al~\cite{gelber2018freeform} presented a complete forward search algorithm for a 3-axis 3D printer that constrains the deformation of the structure but does not address higher-DOF systems. 
Choreo was the first extrusion planning system that uses  a robot manipulator to make robotic extrusion planning for arbitrary topologies possible~\cite{Huang2018automated}. 
However, Choreo separates the extrusion planning into separate  sequence and transit planning phases, which can prevent the discovery of a feasible solution in certain cases. Additionally, to make the sequence planning tractable, Choreo requires a user-specified ordering on elements to guide the search algorithm.

In most prior work in robotic assembly, the construction of bar structures has also followed a layer-based approach, simplifying particularly the sequence planning and reachability challenge such that it can be intuitively solved~\cite{apolinarska2016squential}.
Recent projects have demonstrated robotic assembly for more complex geometries~\cite{sondergaard2016topology,helm2017additivetimber}, but motion planning has remained a key challenge, addressed by pre-defining the robot’s path through a trial and error process. 
To construct large-scale differentiated space-frame structures, recent research has focused on integrating motion planning and sequence definition into the design process. 
One approach to addressing the duality of sequence- and motion-planning is to define the sequence of a spatial assembly simultaneously with the geometry generation, within the design process. 
Parascho et al.~\cite{parascho_computational_2018} propose a constructive system relying on tetrahedral configurations, which, if assembled in the order that the bars are generated, provides local structural stability and support during construction. While this method removes the sequencing challenge from the fabrication process, it still relies on motion planning algorithms to identify collision-free trajectories, which are not guaranteed to be readily found. In order to find such trajectories, a trial-and-error process was needed to iterate through multiple design possibilities that alter both the sequence and the geometry, making the design and planning process very time-consuming.

In the robotic planning literature, task and motion planning (TAMP) involves planning both high-level objectives as well as low-level robot motions required to complete a multi-step manipulation~\cite{garrettIJRR2018,srivastava2014combined,toussaint2015logic}.
For extrusion and assembly problems, the high-level decisions involve the construction sequence, and the low-level motions are the trajectories for motion primitives, like extrusion, pick, place, transit and transfer. 
Although TAMP includes a much broader class of problems than extrusion and assembly for construction, a key challenge in extrusion and assembly problems is that they often require manipulating many more objects, which, as a result, leads to substantially longer planning horizons. 
Solutions to most TAMP benchmarks involve fewer than 30 high-level actions while many construction problems may require assembling several hundreds of elements~\cite{lagriffoulbenchmarks}. Readers are referred to Garrett et al.~\cite{Garrett2021} for an extensive review of the work in this area.

Some notable works consider construction sequencing with a structural stability constraint but in the absence of a robot manipulator. Beyeler et al.~\cite{beyeler2015graph} proposed a backward search algorithm to find stable deconstruction sequences of masonry structures. Deuss et al.~\cite{deuss2014assembling} presented a divide-and-conquer algorithm to compute cable-assisted, self-supporting construction sequences for masonry structures. 
Connecting such approaches with robotic assembly constraints is more challenging algorithmically but necessary to support automated robotic assembly planning. 
The research goal of this paper’s work is to build upon existing material systems and knowledge from the architectural robotics literature and algorithmic insights from the robotic planning literature, to develop generalized planning algorithms that automate the tedious task of programming robots to additively construct various bar structures.


\section{Robotic additive construction}\label{sec:additive}

In this work, we provide a unified formulation of robot-enabled additive construction. Although we focus on spatially extruded systems and bar assembled systems, our framework can be applied to other additive construction tasks if an appropriate set of primitives are specified. 
This formulation allows us to describe and solve problems involving different kinds of systems using the same high-level algorithms. 
In this section, we show that, at a high-level, the same constraints are in play during both extrusion and assembly. Because of this, we only need to implement several manipulation-specific motion primitives to plan in different construction settings. We first describe two types of constraints: structural (Section \ref{sec:stiffness}) and robot-involved geometric constraints (Section \ref{sec:geometric}). 
We use the concepts of action templates and plan skeletons as modeling abstractions that represent motion primitives in different applications. 
Then, we develop planning algorithms that are able to find feasible construction and robotic motion plans for any plan skeleton, which in turn results in a plan for the underlying bar structure construction task (Section \ref{sec:planning_construction}).

\subsection{Bar systems}\label{sec:bar_system}

\begin{figure}[htb]
 \centering
 \includegraphics[width=0.48\columnwidth]{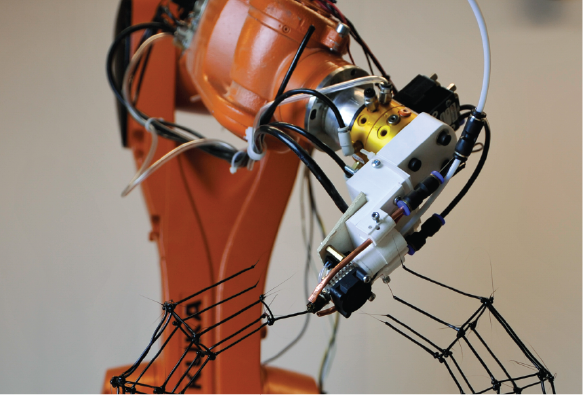}
 \includegraphics[width=0.486\columnwidth]{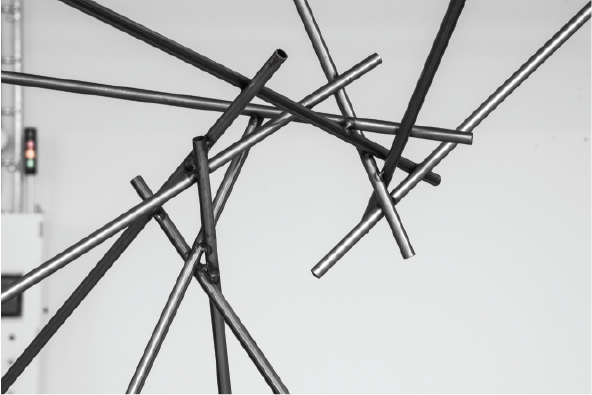}
 \caption{Connection examples between bar elements; 
 {\it left}: two extruded plastic bars are connected exactly at their shared end point \cite{garrett2020scalable}; 
 {\it right}: Point-welded connections between steel bars \cite{parascho_cooperative_2019}.}
 \label{fig:connector}
\end{figure}

A bar system can be described as a collection of $n$ linear elements $E$. Each linear element corresponds to a bar. Each element can be connected to one or more other elements. 
In spatial extrusion, these linear elements are solid plastic cylinders and are connected exactly at the endpoint of the linear element (Fig. \ref{fig:connector}-left). At the connection point, an extra sphere of extruded plastic is applied to form a ball joint-like connection among elements after solidification. 
In bar assembly, elements can be connected anywhere along the boundary of the element (Fig. \ref{fig:connector}-right). In this case, extra adhesive material is applied between elements to form a rigid connection, such as welding metals, glue, etc. 
We assume that the connectors can transfer moment and torsional load to give the partially constructed structure the ability to cantilever, which is a reasonable assumption given the rotational stiffness of these joints. 
We assume that the robot does not exert additional external forces on the structure, and thus each partial structure only experiences forces induced by self-weight. 
Because of this, structural constraints only involve the partial construction, but geometric constraints involve the interaction between the robot and the structure it is building.

For a bar system, we define a {\it construction sequence} to be an ordering of the elements that dictates the sequence for which they added to the structure. In the scope of this work, elements are not allowed to be removed and later re-introduced once they are placed. 
Furthermore, we do not allow extra scaffolding elements to be introduced. We call such a construction process a {\it monotonic} additive construction process because the structure only grows over time.

A construction sequence is {\it valid} if every partial structure along the sequence satisfies both the structural constraints that limit the deformation of the partial construction and the geometric constraints induced by the behavior of the robotic manipulator. Such geometric constraints ensure that the robot stays collision-free with the partial construction, as well as restricting its trajectory to follow certain motion primitives.

In the following sections, we will describe the details of the structural constraints and robotic geometric constraints. We shall see that through the lens of action templates and plan skeletons, we can unify the treatment of both extrusion and assembly planning under the same framework, and reduce their implementation as a switchable module in our general backward state-space search algorithm.

\subsection{Structural constraints}\label{sec:stiffness}

In this paper, we focus on deformation-based stiffness constraints as the key structural criterion. A stiffness constraint requires that a partially constructed structure’s maximal nodal deformation is below a given tolerance. 
Each bar element experiences a self-weight load due to gravity, which causes the structure to bend. Excessive deformation is undesirable since it leads to connection failures for subsequently constructed elements. 
The deformation of all nodes is calculated using a first-order linear elastic finite element analysis (FEA) of a 3D frame structure~\cite{McGuire_Gallagher_Ziemian_1999}, and the norm of the $x,\,y,\,z$ translational deflections at each node is compared to a maximum permitted tolerance. 
For extruded structures, the central axes of the extruded elements are directly used as linear beam elements. For assembled structures, there are two types of structural members, both modeled with linear beam elements: (1) bars which represent the rods and (2) connectors which represent the welded or glued connection between two bars~\cite{parascho_cooperative_2019}.

\subsection{Geometric constraints}\label{sec:geometric}

The second key type of constraint involves the geometric interaction between the robot, the environment, and each partially constructed structure. 
First, the robot must always respect common motion constraints such as staying within its joint limits as well as avoiding collisions with itself, the environment, and the currently assembled elements. 
In addition, different types of motion primitives impose additional task-space constraints on the robot. In this section, we will examine several motion primitives and discuss the constraints they induce in extrusion and assembly problems.

We introduce the following terminology to discuss different motion primitives using the same language. 
Let {\it action template} be a parameterized robot skill that describes a type of robot motion. Additionally, let a {\it plan skeleton} be a sequence of action templates~\cite{Garrett2021}. For extrusion and assembly, each plan has a corresponding plan skeleton that follows a known repetitive pattern and has a known fixed length. Through formalizing the robot’s skill primitives and corresponding action templates, we identify constraints for different primitives and unify our algorithmic treatment of both extrusion and assembly planning.

\begin{figure}[htb]
 \centering
 \includegraphics[width=1\columnwidth]{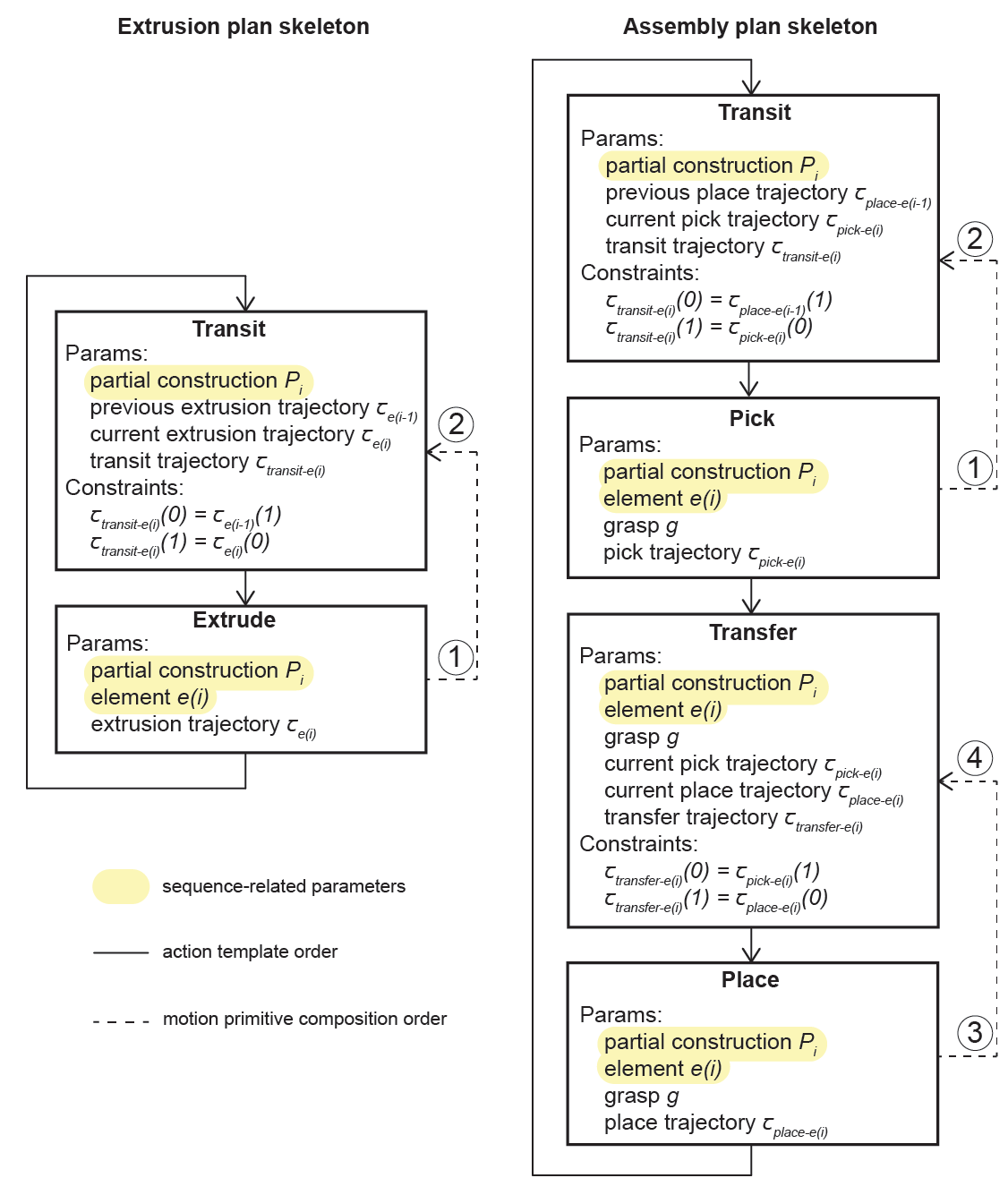}
 \caption{
 The plan skeleton for extrusion ({\it left}) and assembly ({\it right}). On each iteration $i \in \{1, ..., n\}$ of the plan skeleton, a new element $e(i) \in E$  is constructed where $e(i)$ is selected by a planner.
 }
 \label{fig:plan_skeleton}
\end{figure}

In an extrusion process, the robot only has two skill primitives: extrude and transit. The robot alternates between performing extrude primitives, where the robot is extruding material, and transit primitives, where the robot is moving to another element instead of extruding. The plan skeleton for extrusion is an alternating sequence of the transit and extrude templates (Fig. \ref{fig:plan_skeleton} -left). Detailed descriptions of the constraints involved in these action templates and the primitive planners involved can be found in Garrett et al.~\cite{garrett2020scalable}.

In an assembly process, the robot repetitively performs the following four skill primitives in order: (1) {\it transit}, (2) {\it pick}, (3) {\it transfer}, and (4) {\it place}. 
Unlike extrusion, where materials are distributed directly out of the hotend end effector, assembly involves re-positioning ready-made elements. 
Typically, the elements are initially placed on a material rack and must be picked up; however, in some applications, a human operator instead manually lifts and positions new elements in the robot’s gripper. The pick and place primitives attach and detach the element, and the transit and transfer primitives move between each pick and place. 
The pick and place primitives both have two sub-procedures, approach and retreat, which move the robot in and out of contact with the element being manipulated.

\begin{figure}[h!]
 \centering
 \includegraphics[width=1\columnwidth]{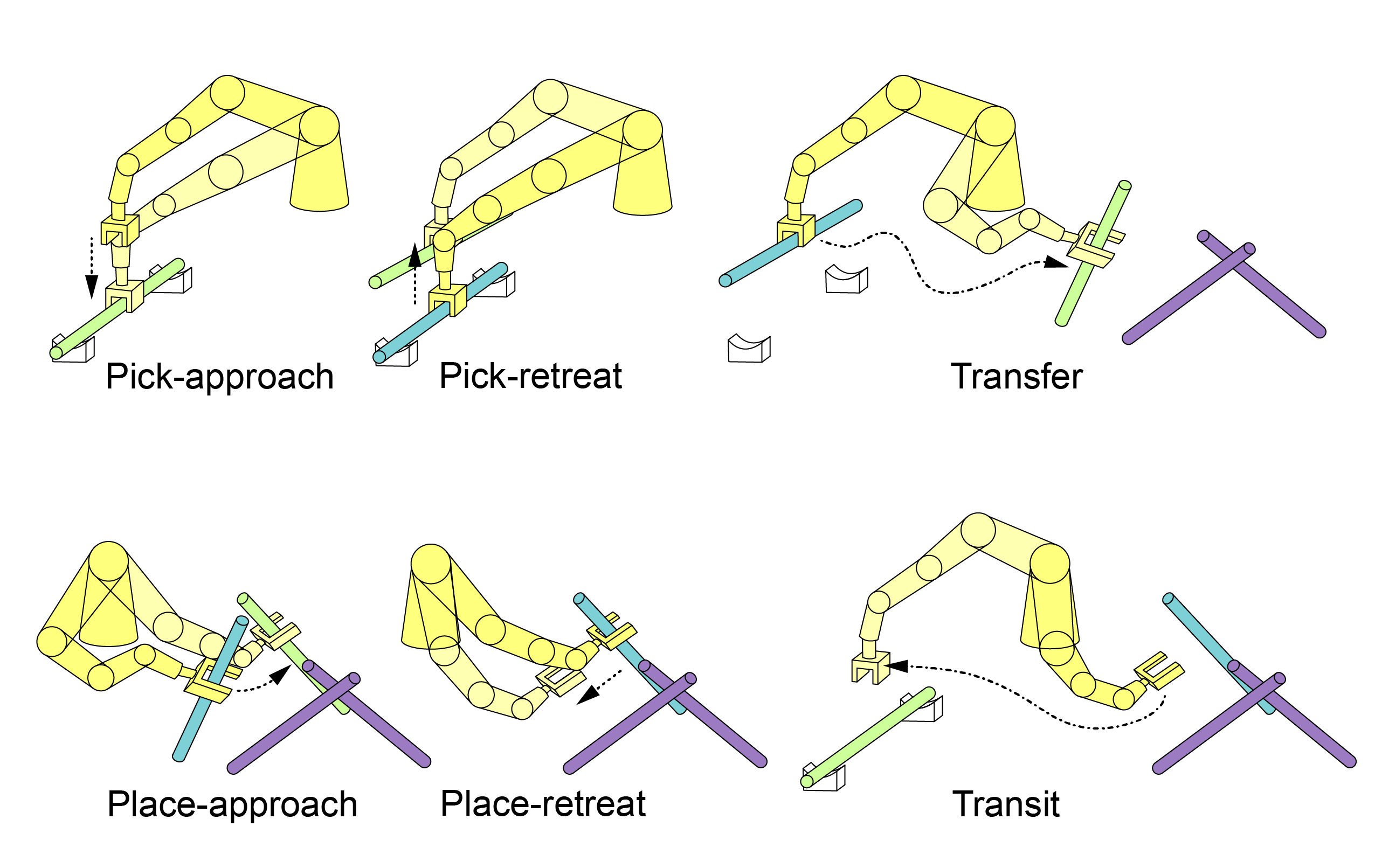}
 \caption{Motion primitives for assembling a single bar element.}
 \label{fig:assembly_mode}
\end{figure}

In a pick primitive, the robot approaches an element on the material rack ({\it pick-approach} in Fig. \ref{fig:assembly_mode}), grasps it, and moves it out of contact with the rack ({\it pick-retreat} in Fig. \ref{fig:assembly_mode}). 
The parameters for the pick action template include: (1) the partial structure $P$ that specifies the elements that the robot must not collide with, (2) the element $e$ being picked up, (3) a grasp pose $g$ that describes the relative transformation between the robot’s end effector and the element when attached, and (4) a pick trajectory $\tau_{\textit{pick}-e}$, which is the concatenation of the pick-approach and pick-retreat trajectories.

In a place primitive, the robot inserts the element into the structure ({\it place-approach} in Fig. \ref{fig:assembly_mode}), detaches it, and moves out of contact with the structure ({\it place-retreat} in Fig. \ref{fig:assembly_mode}). Because the held element during an insertion motion is in the close vicinity of the assembled elements, the place-approach motion usually requires the path of the grasped element to involve both translation and rotation. 
In contrast, the place-retreat motion is often just a translational motion that moves the end effector out of contact with the element after detachment. 
With the exception of the place trajectory parameter $\tau_{\textit{place}-e}$, the place action template shares the same parameter values as its corresponding pick action template, including the partial structure $P$, the element $e$, and the grasp pose $g$.

The robot moves between two configurations using the transit and transfer primitives. In a transit primitive, the robot’s end effector is empty. 
However, in a transfer primitive, the robot is grasping an element at the relative pose specified by the grasp pose parameter $g$ of the adjacent pick and place action templates. Transfer primitives impose two additional constraints that (1) the grasped element remains rigidly attached to the robot’s end effector and (2) the grasped element does not collide with the robot nor the currently built structure. 
As a result, the grasped element can be seen as a temporary component of the robot, and because it only increases the volume occupied by the robot, it decreases the collision-free configuration space of the robot. 
The transit action template’s parameters include (1) the partial structure $P$, (2) the place trajectory $\tau_{\textit{place}-e'}$ from the previous construction step, 
(3) the pick trajectory $\tau_{\textit{pick}-e}$ from the current construction step, and (4) a transit trajectory $\tau_{\textit{transit}-e}$. The transfer action template’s parameters are (1) the partial structure $P$, (2) the grasped element $e$, (3) the grasp pose $g$, (4) the current pick trajectory $\tau_{\textit{pick}-e}$, (5) the current place trajectory $\tau_{\textit{place}-e}$, and (6) a transfer trajectory $\tau_{\textit{transfer}-e}$.
 
The plan skeleton for assembly is a repeating sequence of the four action templates in order: transit, pick, transfer, and place (Fig. \ref{fig:plan_skeleton}-right). Within each iteration, a planner must select the partial construction $P$, assembled element $e$, and grasp pose $g$ as well as valid robot trajectories. 
Similar to extrusion, because of parameter dependence, although the transit and transfer action templates come {\it before} pick and place in the plan skeleton, their parameters are determined {\it later} during planning.

Transit and transfer primitives can be implemented using any standard motion planner, such as the rapidly-exploring random tree (RRT) algorithm~\cite{lavalle1998rapidly}. Additionally, any constrained motion planner can be used to find collision-free paths for the pick and place primitives~\cite{berenson2011task,kingston2019exploring,stilman2010global}.


\section{Planning for robotic construction}\label{sec:planning_construction}

In Section \ref{sec:additive}, we observed that both extrusion and assembly require a planner to (1) find a structurally feasible construction sequence for the bar elements and (2) compute robotic motions for the corresponding action primitives while constructing each element. 
In this section, we will describe state-space planning algorithms for additive robotic construction. Additionally, we will describe why searching backward helps the algorithm avoid geometric deadends, which improves the planner’s performance. Finally, we will explain the impact of different search heuristics on a planner’s performance.

One common planning strategy is to hierarchically decompose (1) selecting the construction sequence and (2) planning robot motions into two separate sequential planning problems~\cite{Huang2018automated}. 
In this strategy, the robot is not considered at all when planning the construction sequence in step (1), and the construction sequence is fixed when planning robot motions in step (2). 
However, excluding geometric constraints from the construction sequence planning can often lead to construction sequences that are infeasible for a robot to construct, causing the planner to fail~\cite{Huang2018automated}. Instead, by jointly planning the construction sequence and robot motions, we can develop planning algorithms that are complete, {\it i.e.} are guaranteed to find a correct solution if one exists.

\subsection{State-space planning algorithms}\label{sec:state_space_search}

State-space planning algorithms search over possible sequences of world states. Here, a state represents the status of a step in the construction process. 
A state includes (1) the partially constructed structure (2) the robot’s current configuration, and (3) any values pertaining to active constraints imposed by the current motion primitive, such as the constant orientation of the end effector for extrude motions and the rigid grasp pose of an element for transfer motions. 
Because both the {\it discrete} constructed structure and the {\it continuous} robot configuration are simultaneously parts of the state, the state is {\it hybrid}. Furthermore, the set of possible states is infinitely large, whereas the set of states when only considering sequencing is finite.

A state-space planning algorithm can either search {\it forward} in time starting from the initial state or {\it backward} in time from a goal state. 
In the context of construction, a forward search starts from the unbuilt structure and sequentially adds elements. Thus, it can be seen as exploring different ways of {\it constructing} the structure. In contrast, a backward search starts from the completed structure and sequentially removes elements, thus {\it deconstructing} the structure. We discuss the trade-offs when moving forward versus backward in Section \ref{sec:backward}.


\subsection{Backward search}\label{sec:backward}

Forward search is the most intuitive way to search because it mimics how plans will be executed in the real world. However, our previous work on robotic extrusion planning found that forward search encounters many {\it dead ends}, states that do not lead to any plan, due to geometric constraints~\cite{garrett2020scalable}. 
For example, a state in the forward search might have the outer layer of the structure constructed with the interior remaining. Because the outer layer prevents the robot from reaching any element in the interior without colliding, the search cannot progress from this state. 
However, in a backward search, elements are removed instead of added, so the search could instead peel-off the exterior in order to reach the interior. 
As a result, we found that searching backward can significantly reduce the number of states explored to find a plan and thus dramatically reduce the computation time.

\begin{figure}[htb]
 \centering
 \includegraphics[width=1\columnwidth]{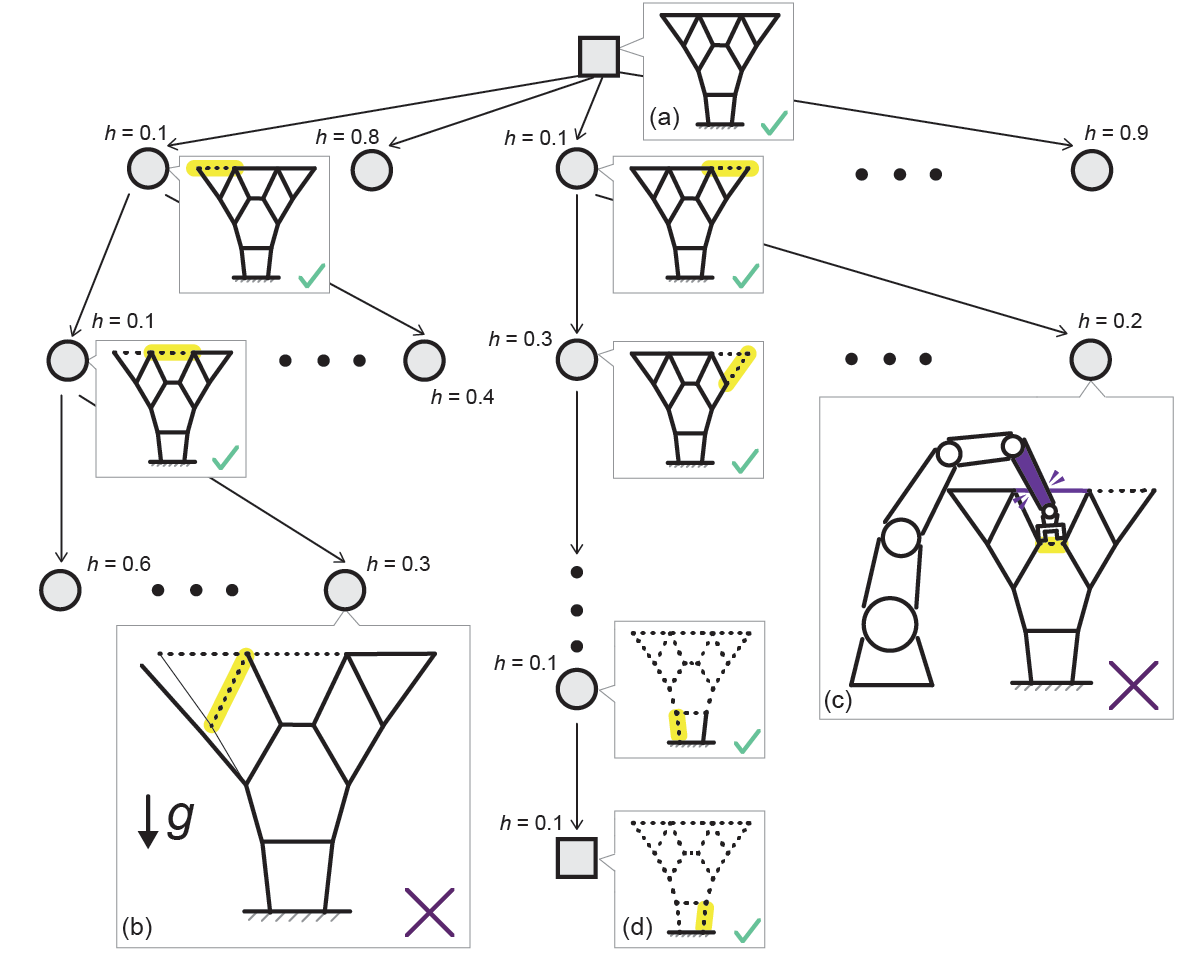}
 \caption{An example search tree for our backward state-space search algorithm.
 }
 \label{fig:state_space_graph}
\end{figure}

Fig. \ref{fig:state_space_graph} illustrates an example of how the backward state-space search algorithm operates. The backward search starts from the completed structure state at the top rectangle (Fig. \ref{fig:state_space_graph}-a) and works its way downward towards the initial unbuilt state at the bottom (Fig. \ref{fig:state_space_graph}-d) through intermediate partially constructed states, which are circles. 
The partially constructed structure $P$ is illustrated right next to each state.
Every time the algorithm selects the next element to remove, it must evaluate the new partial construction’s structural feasibility (Fig. \ref{fig:state_space_graph}-b) as well as geometric constructability (Fig. \ref{fig:state_space_graph}-c). 
Specifically, it is geometrically constructable if the motion primitive planners for the action templates in Section \ref{sec:geometric} succeed in finding trajectories. 
To proceed towards the next state, the algorithm must choose an element among the candidate elements that have not yet been removed, according to a ordering induced by the heurstic function $h(e)$ (illustrated as a normalized number $0 \leq h \leq 1$ next to each transition arrow between states in Fig. \ref{fig:state_space_graph}). If there is no feasible successor state, the algorithm encounters a dead end, and it will {\it backtrack} to the previous state, undoing the previous deconstruction action, to explore other options from there.

\begin{figure*}[h!]
 \centering
 \includegraphics[width=0.6\textwidth]{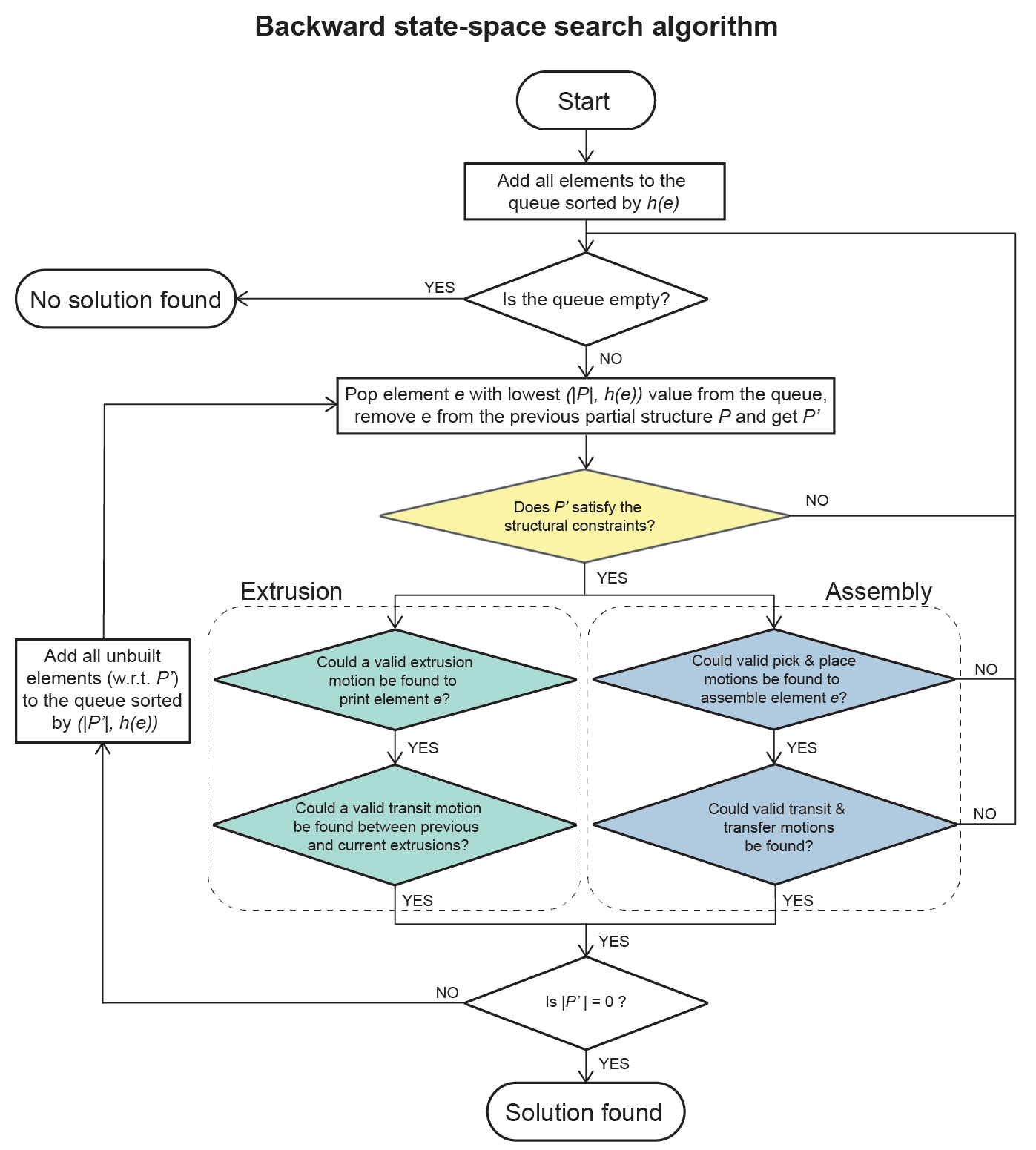}
 \caption{A flowchart describing our backward state-space search algorithm.}
 \label{fig:backward_search_flowchart}
\end{figure*}

We visualize the behavior of the backward search algorithm through the flow chart in Fig. \ref{fig:backward_search_flowchart}. We use a priority queue to sort states according to the pair $(|P|, h(e))$, which consists of the number of remaining elements and a heuristic value $h(e)$ defined on element $e$. 
The priority queue is sorted in an ascending order by first considering element with lower $|P|$ and $h(e)$ values.
Minimizing the number of remaining elements results in a {\it greedy} search that attempts to complete the plan as quickly as possible, akin to a depth-first search. 
At the start of each iteration, the element with the lowest number of remaining elements is popped, where the heuristic value $h(e)$ is a tie-breaker. This ensures that the search always prefers progressing towards removing more elements. The yellow-shaded diamond box in Fig. \ref{fig:backward_search_flowchart} signifies the structural constraint evaluation. The blue- and green-shaded diamond boxes correspond to the geometric constraint evaluations for extrusion and assembly. 
Both constraint evaluations are computationally expensive, because they rely on physical simulation and primitive-specific motion planning. Thus, the total efficiency of the algorithm depends on the number of states that are visited and checked, which can grow enormously if the search encounters dead ends.

Following this principle, the proposed backward search algorithm has its search direction designed to address geometric constraints, and its heuristic function tailored for the structural constraints. For the geometric constraints, removing an element increases the collision-free configuration space of the robot, since there are fewer collision bodies in the workspace. Thus, the robot is most geometrically constrained at the beginning of the backward search, limiting which elements can be initially constructed.

Although backward search makes geometric planning easier, it comes at the expense of making the structural constraints more difficult to satisfy. At the beginning of a backward search, there are many elements that can be prematurely removed without violating the structural constraints. However, as the search progresses, there are fewer options for supporting the structure, making the search more likely to reach a structural dead end. Thus, we propose a heuristic function that can bias the search to adhere to a structurally feasible construction sequence as much as possible while still satisfying the robot-related geometric constraints.

One intuitive heuristic is to remove elements based on their distance to the ground, where elements farther away from the ground are removed first. This heuristic is motivated by human builders’ common construction practices. When building self-supporting structures, builders typically construct elements in a layer-by-layer fashion, in an order opposing gravity. This strategy originates from the physical reasoning that elements that are lower in the layer contribute to support the elements above them. Such heuristic functions can be seen as applying "soft" partial ordering constraints that steer the search. If the heuristic guides the search into a dead end, the algorithm will simply backtrack to previous states, thus preserving completeness. This is in contrast to the hard partial-ordering constraints used in prior work~\cite{huang2016framefab,wu2016printing,Huang2018automated}.

Although effective in many practical situations, this distance-based heuristic might still lead to unnecessary state explorations in some special cases, such as when a structure is hung from the ceiling. 
To overcome some of these disadvantages, we propose a heuristic that directly takes into account the structural constraints. This heuristic performs a {\it forward} search for a valid construction sequence considering {\it only} structural constraints ({\it i.e.} ignoring the robot) that is then used to bias the overall {\it backward} search to remain as close as possible to this construction sequence. 
More precisely, it precomputes a valid construction sequence $S$ using a greedy forward search and returns the negated index $i$ of element $e$ in the sequence. 
Namely, if $S[i] = e$ then its heuristic value $h(e) = -i$ (recall that an element with {\it lower} heurstic value will be removed first). Previous work has showed that this heuristic ({\it StiffPlan}) outperforms the distance-based heuristic ({\it EuclideanDist})~\cite{garrett2020scalable}.


\section{Case studies}\label{sec:case_studies}

This section presents case studies of applying our planning algorithm to various extrusion and assembly problems. We present results demonstrating assembly planning for a 7-DOF robot system, with an IRB 4600 robot mounted on a 4-meter-long ABB IRBT 4004 linear track (Section \ref{sec:tangent}). 
A summary of these assembly case studies can be found in Table \ref{table:case_study_list}. More extensive studies on extrusion problems can be found in previous publication~\cite{garrett2020scalable}.

We implemented transit and transfer primitives using RRT-Connect~\cite{KuffnerLaValle}. We used IKFast, an analytical inverse kinematic solver, to initialize the extrude, pick, and place primitives~\cite{diankov2010automated} and Randomized Gradient Descent~\cite{stilman2010global,yao2007path} to plan full paths. We use PyBullet~\cite{coumans2015bullet} for collision checking, forward kinematics, and visualization during motion planning. An open-source Python implementation of our algorithm is available at \url{https://github.com/yijiangh/coop_assembly}.

\begin{table*}[]
\centering
\begin{tabular}{|l|l|l|l|l|l|}
\hline
Design name & \# Elements & \begin{tabular}[c]{@{}l@{}} Connection\\type \end{tabular} & \begin{tabular}[c]{@{}l@{}}Deformation\\tolerance (mm)\end{tabular}  & \begin{tabular}[c]{@{}l@{}}Dimensions\\ (m)\end{tabular} & \begin{tabular}[c]{@{}l@{}}Planning\\ time (min)\end{tabular} \\ \hline
\begin{tabular}[c]{@{}l@{}} Short Arch \\ Fig. \ref{fig:sequence_table}-row 1 \end{tabular} & 39 & double-tangent & 5 & 1.1 x 2.5 x 1.4 & 2.6 \\ \hline
\begin{tabular}[c]{@{}l@{}} Tall Arch \\ Fig. \ref{fig:sequence_table}-row 2 \end{tabular}& 42 & double-tangent & 3 & 0.9 x 1.7 x 1.4 & 5.3 \\ \hline
\begin{tabular}[c]{@{}l@{}} Column \\ Fig. \ref{fig:sequence_table}-row 3 \end{tabular} & 39 & double-tangent & 5 & 1.0 x 2.7 x 1.3 & 3.3 \\ \hline
\begin{tabular}[c]{@{}l@{}} Hydra \\ Fig. \ref{fig:sequence_table}-row 4 \end{tabular} & 42 & double-tangent & 5 & 1.0 x 1.8 x 1.5 & 5.2 \\ \hline
\begin{tabular}[c]{@{}l@{}}TopOpt Vault\\ (assembly, Fig. \ref{fig:topopt_extrusion_assembly}-left)\end{tabular} & 68 & ball joint & 3 & 1.0 x 1.0 x 1.0 & 6.7 \\ \hline
\begin{tabular}[c]{@{}l@{}}TopOpt Vault\\ (extrusion, Fig. \ref{fig:topopt_extrusion_assembly}-right)\end{tabular} & 76 & ball joint & 1.5 & 0.35 x 0.35 x 0.35 & 4.4 \\ \hline
\end{tabular}
\caption{Overview of the assembly case studies for our additive construction method.  The planning time results are for a consumer-grade laptop.}
\label{table:case_study_list}
\end{table*}







\subsection{Assembling double tangent bar systems}
\label{sec:tangent}

Spatial structures made of bar elements are particularly efficient solutions for spanning structures. We therefore focused on designs that showcase the advantages of spatial structures through their structural potential. 
Traditionally, space-frames are designed as a layer-based system with a regular grid of diagonals between horizontal elements. By utilizing robotic assembly, we can free up the design of space frames to allow for different geometric configurations that may result in free-form shapes.

Among the six case studies tested in this paper, we chose to physically demonstrate our method with the construction of an arched truss (called {\it Short Arch} here), consisting of 39 elements, to address a multitude of challenges that are inherent to the robotic assembly of spatial structures. 
An arching geometry presents a particularly challenging structural case, due to the different support conditions during and after construction. Traditionally, arches are built from both supports inwards as cantilevering structures (or using additional support) until they are connected in the middle to form a self-stable structure with two supports. Other assembly sequences are possible, but will strongly influence the structural behavior during construction and might lead to either the necessity of temporary supports or over-sizing of the members to ensure structural integrity at every step of assembly. As a result, the assembly sequence plays a particularly crucial role in the stable construction of arching structures.

\begin{figure*}[]
 \centering
 \includegraphics[width=0.45\textwidth]{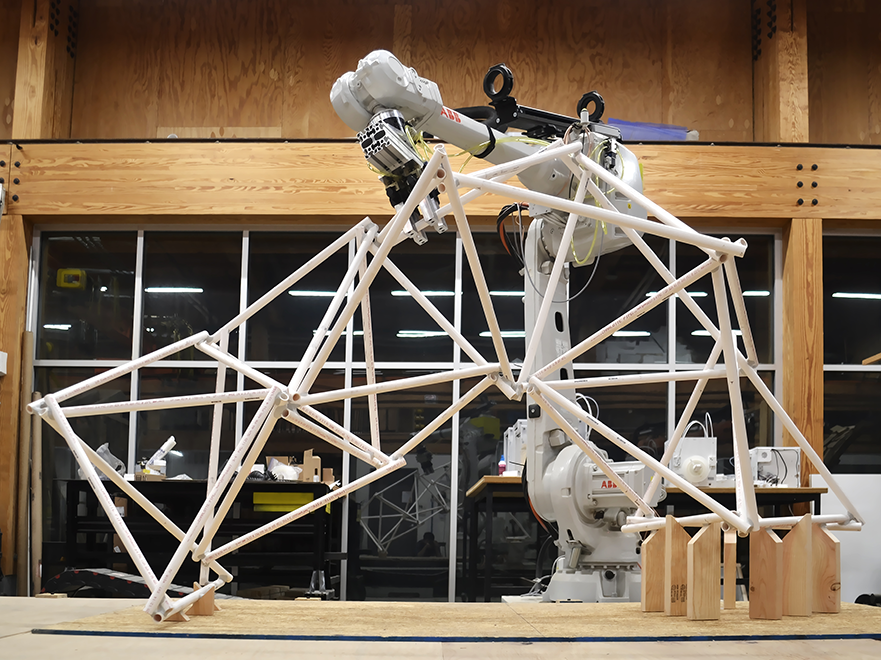}
 \includegraphics[width=0.45\textwidth]{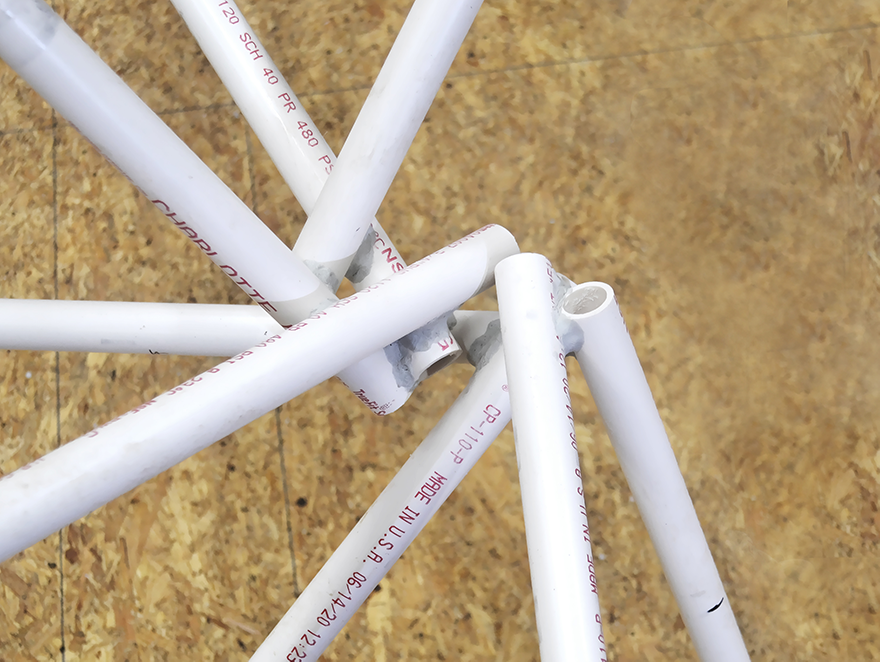}
 \caption{Short Arch; {\it left}: the completed structure; {\it right}: a detailed view of a joint.}
 \label{fig:SP_arch_joint_and_overall}
\end{figure*}

The Short Arch - along with the Tall Arch, Column, and Hydra designs - uses the double-tangent system proposed in Parascho et al.~\cite{parascho_computational_2018} and is based on regular tetrahedra (Fig. \ref{fig:SP_arch_joint_and_overall}). This system presents a number of geometric constraints that need to be fulfilled (each bar needs to be connected to the structure in at least two points at every end) and results in complex nodes where only two elements connect at any single point. The bars’ double tangent connections often necessitate a rotational insertion movement by the robot arm to avoid collisions, making it particularly difficult for the planning of the place motion primitive. As such, we chose to test the proposed construction planning algorithm on a design that combines the sequence challenges resulting from structural stability and robotic reachability.
 
\subsubsection{Short Arch prototype}

If a purely geometric planning process were used ({\it e.g.} robot only, not considering structural behavior), the structure would accordingly have been constructed from one support towards the other, following a pre-defined sequence ensuring local stability of the tetrahedra. 
However, the cantilevering of the entire structure before reaching the second support would result in large deformations and potential connection failures. 
By using our planner, a custom sequence could be found which follows the construction practice of the arch-like structure from both supports. While this sequence is not optimal with regards to structural mechanics, it does provide a compromise that ensures structural stiffness and, at the same time, enables the required complex collision-free insertion trajectories to be found.

\begin{figure*}[]
 \centering
 \includegraphics[width=0.45\textwidth]{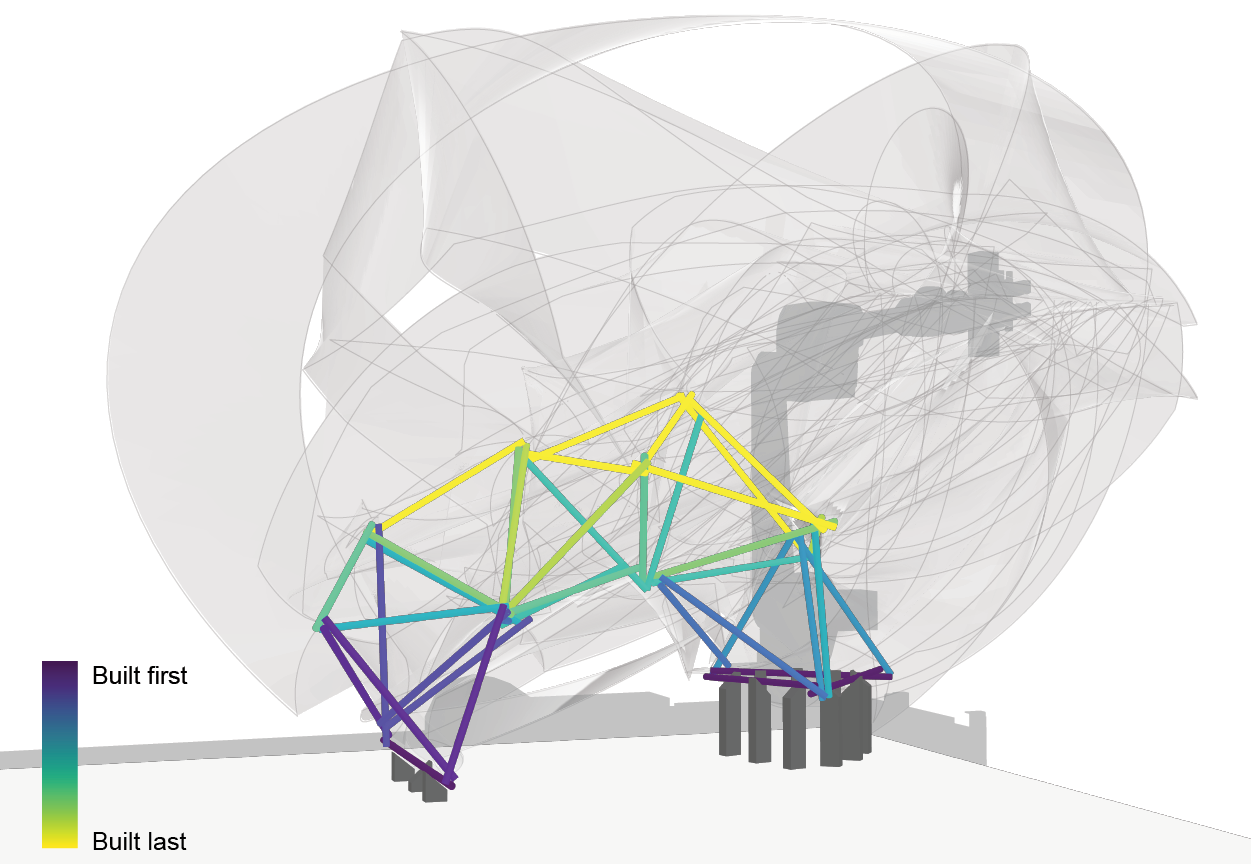}
 \includegraphics[width=0.45\textwidth]{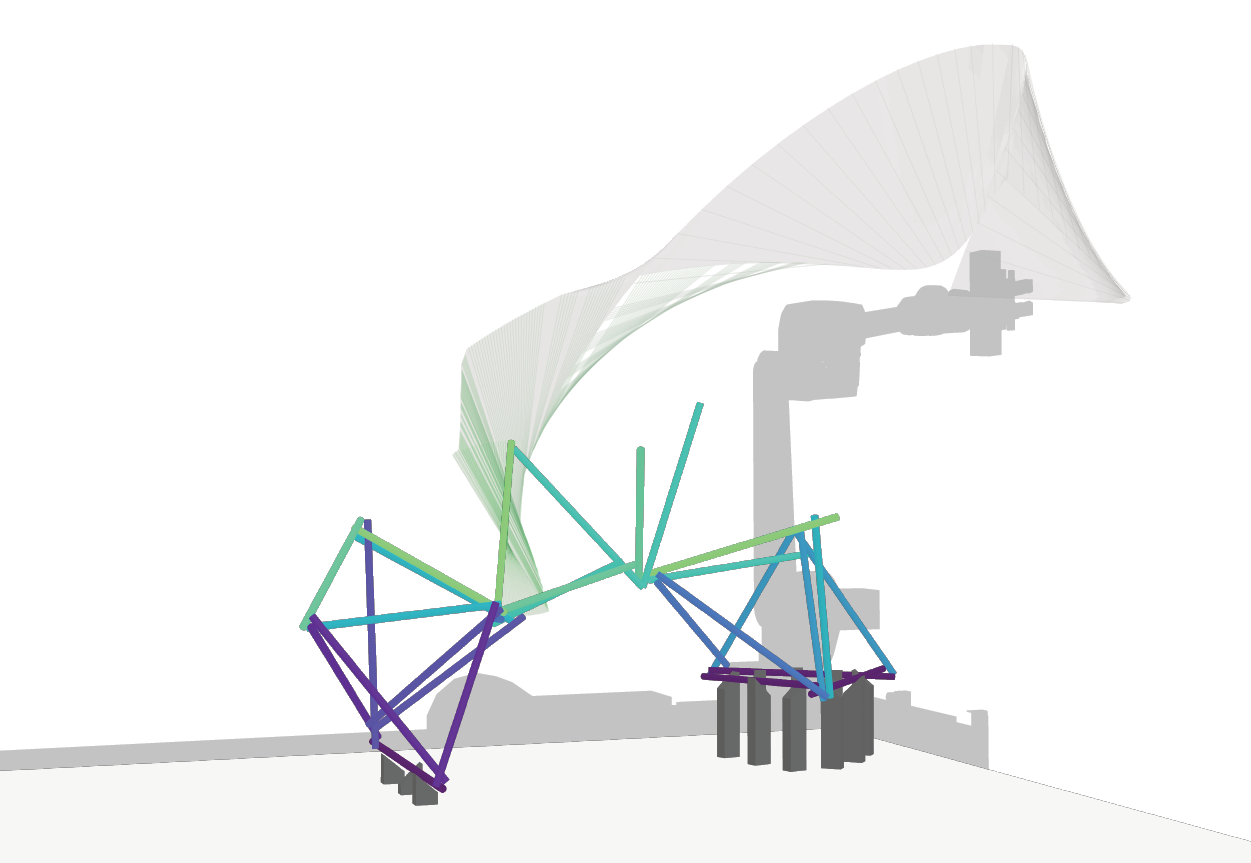}
 \caption{Short Arch; {\it left}: full sequence and trajectories displayed; {\it right}: a transfer and place motion for a particular bar.}
 \label{fig:SP_arch_plan_simulated}
\end{figure*}

The generated construction plan for the Short Arch was tested in a physical prototype using the robotic setup described above (as shown in Fig. \ref{fig:SP_arch_joint_and_overall} and Fig. \ref{fig:SP_arch_plan_simulated}). We constructed the Short Arch structure at a one-to-one-scale using 3/4-inch Schedule 40 (26-mm diameter) PVC pipes joined by epoxy putty (material descriptions given in Table \ref{table:material_descriptions}). The four bottom-most bars of the structure were affixed to wooden blocks connected to a plywood base, according to the structural modeling of these elements as rigid fixities.

\begin{table}[]
\centering
\begin{tabular}{|l|l|}
\hline
Material    & Details                                                                                             \\ \hline
PVC Pipe    & \begin{tabular}[c]{@{}l@{}}Diameter: 26 mm\\ Wall Thickness: 2.9mm\\ Length: 629-945mm\end{tabular} \\ \hline
Epoxy Putty & Oatey Fix-it Stick                                                                                  \\ \hline
\end{tabular}
\caption{The materials used in the physical prototype of arched truss.}
\label{table:material_descriptions}
\end{table}

Epoxy putty was chosen as an adhesive due to its relatively quick setting time (\~20 minutes) and its high viscosity, allowing it to serve as a filler material for the tangent connections between elements and to bridge gaps caused by any combination of inaccuracies due to tolerances between the simulated and physical robot, as well as material deformation. The physical assembly process revealed the expected displacements during construction that are accounted for in our planning method’s FEA simulation: the entire structure deflects upon release of each new element by the robot after joining to the structure. While this was predicted by our planning process, the amount of deflection was often different than predicted, likely due to a combination of inaccuracies of the robotic setup and material modeling. We observed that in many cases, this effect was self-reinforcing: larger than expected deflections caused larger than expected gaps in subsequent bars, requiring larger epoxy joints that therefore deflected more.  This geometric nonlinearity of the deformation was not captured in the FEA simulations, which were based on linear elastic assumptions. Future implementations of this approach could replace the linear elastic FEA with a nonlinear solver.

\begin{figure*}[]
 \centering
 \includegraphics[width=1\textwidth]{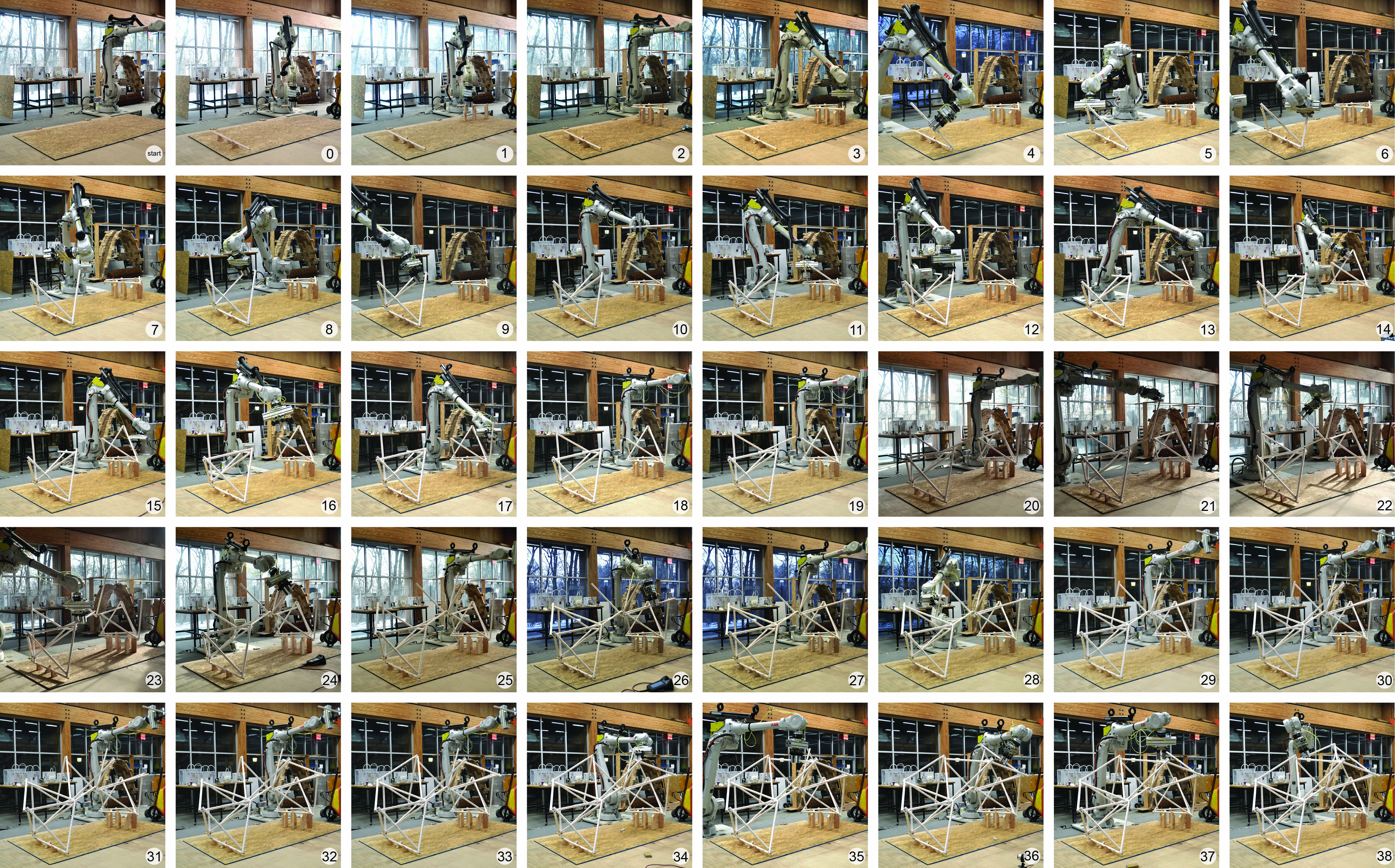}
 \caption{Construction timelapse collage for the Short Arch.}
 \label{fig:SP_arch_collage}
\end{figure*}

\begin{figure}[htb]
 \centering
 \includegraphics[width=1\columnwidth]{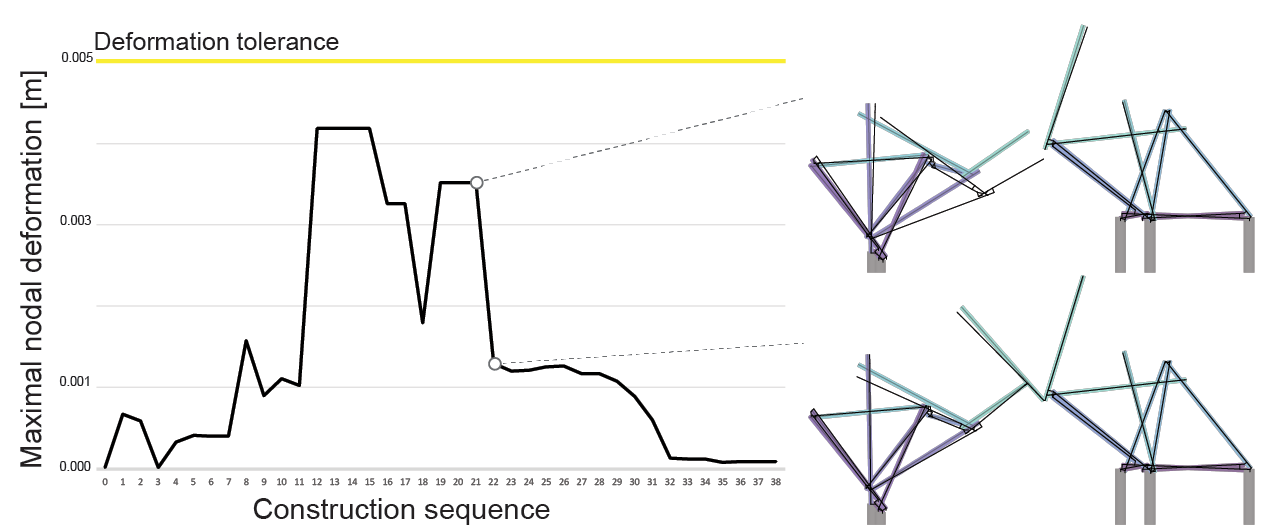}
 \caption{FEA-simulated deformation history of the in-progress structure over the construction sequence of the short arch. The two called out steps show the deformation reduces once a bridging element is placed between the two sides of the construction. The deformation in the called-out steps is magnified 50 times for clarity.}
 \label{fig:deformation_history}
\end{figure}

The maximal deformation tolerance for checking the structural constraint was set in the planning algorithm to be 5mm, though in the majority of cases the real deformation was within the range of 30mm. A larger deformation of approximately 30mm occurred between the placement of bars 19 and 20 of the construction sequence (see frames 19-20 in Fig. \ref{fig:SP_arch_collage}). 
Although these were the steps predicted to have a relatively large deformation (Fig. \ref{fig:deformation_history}), structural joint deformations caused a more dramatic displacement. 
This discrepancy between our structural simulation and the real behavior during construction can be attributed to material modeling inaccuracy, robotic imprecision, and the interrelated compounding effect described above. Additional sources of error may include the observed material creep of the epoxy putty, which led to large deformations overnight after the initial curing of the material. Finally, the spring and joint stiffnesses of the bar-putty interfaces were likely not modeled with complete accuracy.

Despite this unexpected behavior, the resulting misaligned joints were able to be fixed in place using additional epoxy putty, and the spatial positioning of each bar by the robot helped to prevent a catastrophic compounding of misalignments in the overall structure. Further refinement of material stiffnesses in the structural simulation and attention to robotic precision would improve the fidelity of the predicted displacements.

Construction plans for the other three double tangent bar case studies were also generated using our algorithm (although they were not built), as shown in Fig. \ref{fig:sequence_table}. See supplementary materials\footnote{\url{https://youtube.com/playlist?list=PLZhY1DwlPIx6iAOnYxApMZxA-OZieX7q-}} for timelapse videos of the real construction experiment. Note that in the video, the place-approach motions were executed twice, with the first attempt as a dry-run to help identify the positions to apply the epoxy putty.

\begin{figure*}[]
 \centering
 \includegraphics[width=1\textwidth]{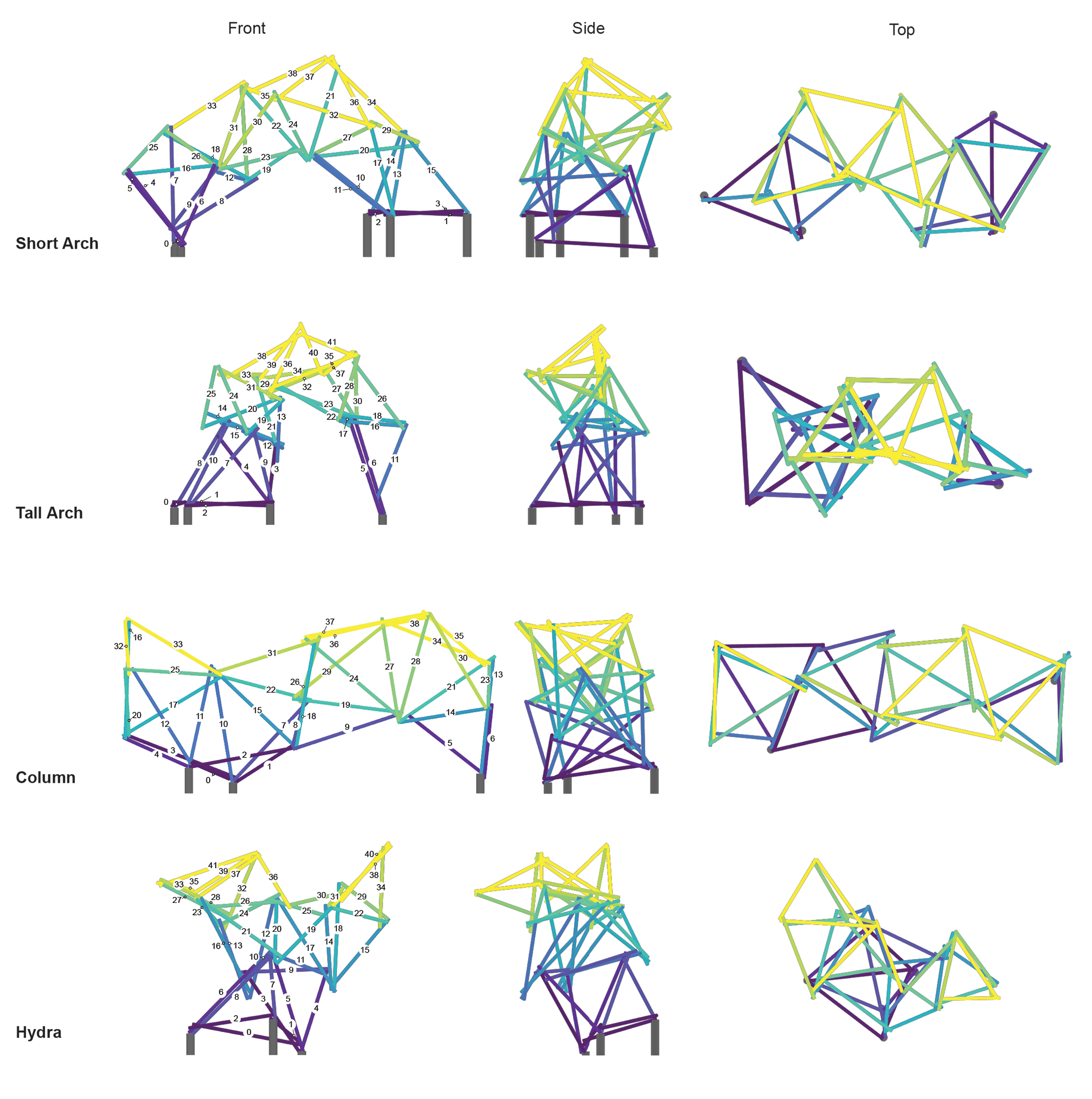}
 \caption{Planned construction sequences (indicated by labeled steps) for the four double tangent assembly case studies.}
 \label{fig:sequence_table}
\end{figure*}

\subsection{Topology-optimized vault}\label{sec:topopt_vault}

\begin{figure*}[htb]
 \centering
 \includegraphics[width=0.48\textwidth]{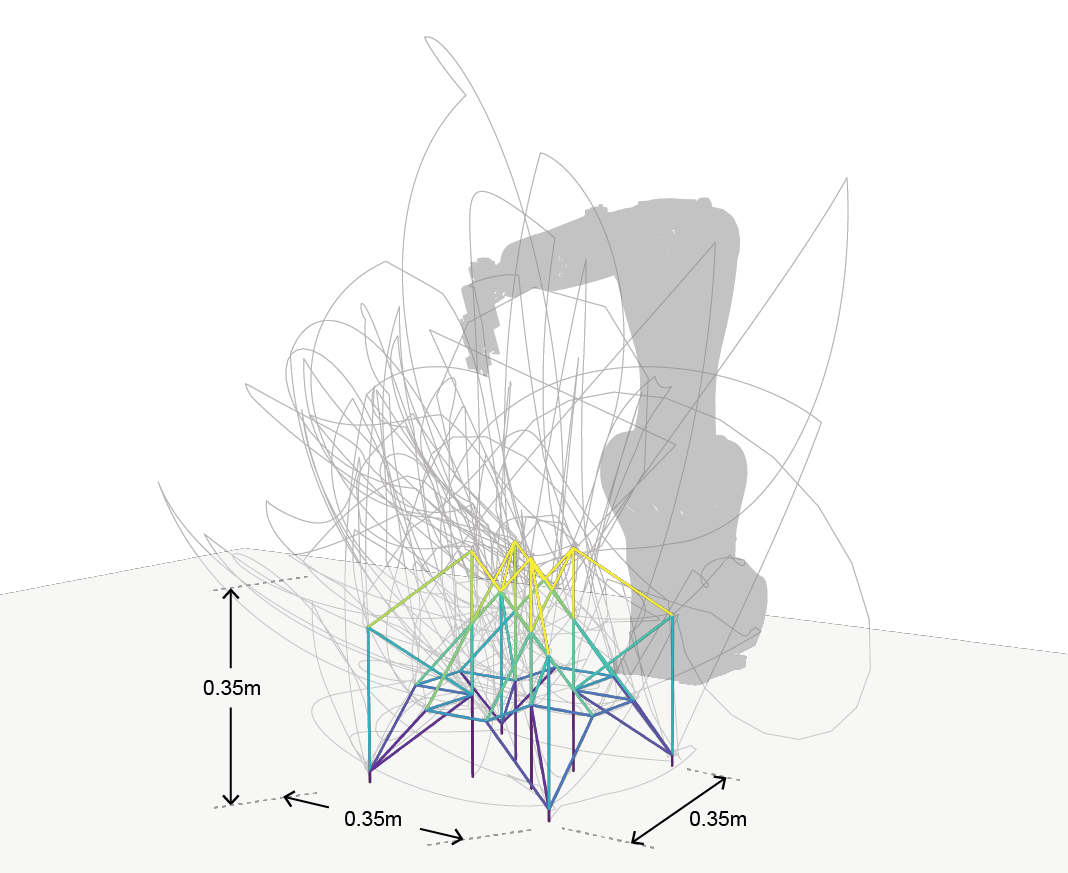}
 \includegraphics[width=0.48\textwidth]{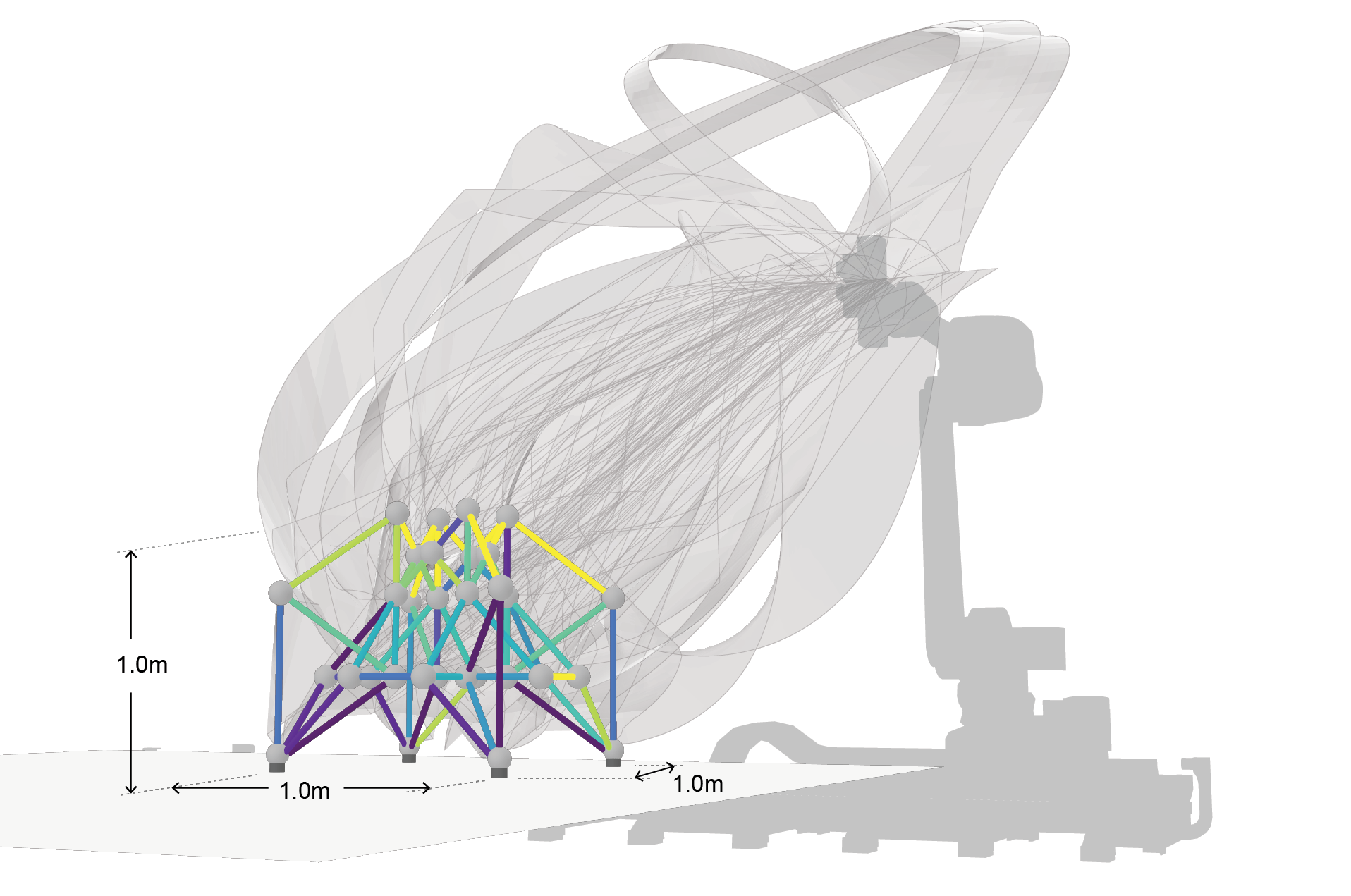}
 \caption{Planned trajectories for extrusion {\em left} and assembly {\em right} of the same design at different scales.}
 \label{fig:topopt_extrusion_assembly}
\end{figure*}

Finally, to further demonstrate the benefits of a general-purpose robotic planner for additive construction, we show planning results of extrusion and assembly of the same design, a topology optimized vault structure, constructed with different materials at different scales (Fig. \ref{fig:topopt_extrusion_assembly}), presented in a previous paper~\cite{garrett2020scalable} in detail for the extrusion case (see Table \ref{table:case_study_list} for key data). It is noteworthy that the construction sequences for the two versions are quite similar (but not identical), but the motion paths are completely different. In the extrusion version, the collision-free space is slightly larger, since the robot is not holding the next bar to be placed. Additionally, the material is lighter, making it easier to meet the structural constraints. The difference in motion is also due to the difference in scales and motion primitives between extrusion and bar assembly.

\section{Conclusions, potential impact, and future work}

This paper has presented a newly unified approach for planning additive construction of bar structures that are made by either spatial extrusion or discrete assembly. By extending previous research, we have leveraged a rigorous state-space search algorithm to jointly find feasible construction sequences and motion plans automatically. In order to efficiently plan in the presence of both geometric and structural constraints, the algorithm performs a backward search to find geometrically feasible trajectories and uses forward reasoning as a heuristic to guide the search through structurally-sound states. We have demonstrated the effectiveness and versatility of this approach on six diverse case studies, five of which are newly presented in this paper and involve bar assembly construction methods. A human-scale physical prototype further tests the method, with a successful result built using the sequence and motion plan generated by our algorithm.

One key advantage of our planning tool is the potential for adaptability to new construction processes, demonstrated here by the application on both extrusion and assembly, thanks to the flexible plan skeleton formulation. With the growing popularity and accessibility of digital fabrication, we are witnessing the development of new methods for construction at an ever-growing pace. In order to propose a relevant planning tool for these processes, it is crucial to consider flexibility with regard to material processes and robotic tasks. We aim to address this by generalizing our approach and making the adjustment to new fabrication methods as easy as possible. Our approach can be extended to new robotic construction systems simply by defining the plan skeleton and associated action templates. In the future, we imagine being able to provide similar sequence and motion-planning tools for applications with multiple machines, or those beyond additive processes, such as material manipulation, disassembly, or reconfiguration. 

The automated nature of our approach also suggests directions for future work regarding design methods. In its current iteration, the method leads to an intuitive design setup, in which a designer does not need to consider fabrication constraints or stability during construction, but can input any design and check its feasibility through the proposed algorithm. However, in developing the work presented in this paper, we sometimes encountered structures with configurations that were difficult, if not impossible, to build. This was often due to the bars not being reachable at all by the robot, which can be a consequence of dimensions, position and orientation of the bars. While our method simplifies the construction of buildable structures (and quickly identifies whether structures are buildable by given robotic setup) we believe that the next step is to achieve a closer integration of the sequence and motion planning into the design process that currently precedes these calculations. Rather than discussing the hierarchical position of the sequence planning - either as a part of or after the design process - we imagine it fully incorporated into the design process itself. The current algorithm runs in minutes on a consumer-grade laptop, which suggests the possibility for real-time construction feasibility feedback for designers. This could be harnessed to avoid post-processing problems and geometric deadends, either in manual design methods, optimization-assisted, or generative methods. For example, we imagine that our method can be incorporated into sequence-based design processes, by providing feedback on motion planning and structural stability at every construction step. This would expand our method by adding the possibility of using the sequence to inform the geometry, as an alternative to using the geometry to define the sequence.










\begin{acknowledgements}
Caelan Garrett acknowledges the support
from NSF grants 1420316, 1523767 and 1723381, from AFOSR
FA9550-17-1-0165, from ONR grant N00014-14-1-0486, and an NSF
GRFP fellow-ship with primary award number 1122374. Any opinions,
findings, and conclusions or recommendations expressed in this material
are those of the authors and do not necessarily reflect the views
of the sponsors.
\end{acknowledgements}

%
%


\begin{thebibliography}{10}
\providecommand{\url}[1]{{#1}}
\providecommand{\urlprefix}{URL }
\expandafter\ifx\csname urlstyle\endcsname\relax
  \providecommand{\doi}[1]{DOI~\discretionary{}{}{}#1}\else
  \providecommand{\doi}{DOI~\discretionary{}{}{}\begingroup
  \urlstyle{rm}\Url}\fi

\bibitem{apolinarska2016squential}
Apolinarska, A.A., Knauss, M., Gramazio, F., Kohler, M.: The sequential roof.
\newblock In: Advancing Wood Architecture: A Computational Approach, pp. 45 --
  57. Routledge (2016)

\bibitem{berenson2011task}
Berenson, D., Srinivasa, S., Kuffner, J.: {Task space regions: A framework for
  pose-constrained manipulation planning}.
\newblock The International Journal of Robotics Research \textbf{30}(12),
  1435--1460 (2011)

\bibitem{beyeler2015graph}
Beyeler, L., Bazin, J.C., Whiting, E.: A graph-based approach for discovery of
  stable deconstruction sequences.
\newblock In: Advances in Architectural Geometry 2014, pp. 145--157. Springer
  (2015)

\bibitem{braumann2011parametric}
Braumann, J., Brell-Cokcan, S.: {Parametric robot control: integrated CAD/CAM
  for architectural design}.
\newblock In: Proceedings of the 31st Annual Conference of the Association for
  Computer Aided Design in Architecture (ACADIA) (2011)

\bibitem{coumans2015bullet}
Coumans, E.: {Bullet physics simulation}.
\newblock In: ACM SIGGRAPH 2015 Courses, p.~7. ACM (2015)

\bibitem{deuss2014assembling}
Deuss, M., Panozzo, D., Whiting, E., Liu, Y., Block, P., Sorkine-Hornung, O.,
  Pauly, M.: {Assembling self-supporting structures}.
\newblock ACM Transactions on Graphics (TOG) \textbf{33}(6), 214 (2014)

\bibitem{diankov2010automated}
Diankov, R.: {Automated construction of robotic manipulation programs}.
\newblock Ph.D. thesis, Robotics Institute, Carnegie Mellon University (2010)

\bibitem{eversmann2017prefabtimber}
Eversmann, P., Gramazio, F., Kohler, M.: {Robotic prefabrication of timber
  structures: towards automated large-scale spatial assembly}.
\newblock Construction Robotics \textbf{1}(1-4), 49--60 (2017)

\bibitem{gandia2018towards}
Gandia, A., Parascho, S., Rust, R., Casas, G., Gramazio, F., Kohler, M.:
  {Towards Automatic Path Planning for Robotically Assembled Spatial
  Structures}.
\newblock In: Robotic Fabrication in Architecture, Art and Design, pp. 59--73.
  Springer (2018)

\bibitem{garrett2020scalable}
Garrett, C., Huang, Y., Lozano-P{\'{e}}rez, T., Mueller, C.: Scalable and
  probabilistically complete planning for robotic spatial extrusion.
\newblock In: Robotics: Science and Systems XVI. Robotics: Science and Systems
  Foundation (2020).
\newblock \doi{10.15607/RSS.2020.XVI.092}

\bibitem{Garrett2021}
Garrett, C.R., Chitnis, R., Holladay, R., Kim, B., Silver, T., Kaelbling, L.P.,
  Lozano-P{\'{e}}rez, T.: {Integrated Task and Motion Planning}.
\newblock Annual review of control, robotics, and autonomous systems \textbf{4}
  (2021)

\bibitem{garrettIJRR2018}
Garrett, C.R.C., Lozano-P{\'{e}}rez, T., Kaelbling, L.L.P.: {Sampling-based
  methods for factored task and motion planning}.
\newblock The International Journal of Robotics Research \textbf{37}(13-14)
  (2018).
\newblock \doi{10.1177/0278364918802962}

\bibitem{gelber2018freeform}
Gelber, M.K., Hurst, G., Bhargava, R.: Freeform assembly planning.
\newblock IEEE Transactions on Automation Science and Engineering
  \textbf{16}(3), 1315--1329 (2018)

\bibitem{gramazio2014robotic}
Gramazio, F., Matthias, K., Willmann, J.: {The robotic touch}.
\newblock Park Books (2014)

\bibitem{hack2014mesh}
Hack, N., Lauer, W.V.: {Mesh-Mould: Robotically Fabricated Spatial Meshes as
  Reinforced Concrete Formwork}.
\newblock Architectural Design \textbf{84}(3), 44--53 (2014)

\bibitem{helm2017additivetimber}
Helm, V., Knauss, M., Kohlhammer, T., Gramazio, F., Kohler, M.: Additive
  robotic fabrication of complex timber structures.
\newblock In: Advancing Wood Architecture: A Computational Approach, pp. 29 --
  43. Routledge (2017)

\bibitem{helm2015iridescence}
Helm, V., Willmann, J., Thoma, A., Pi{\v{s}}korec, L., Hack, N., Gramazio, F.,
  Kohler, M.: {Iridescence print: Robotically printed lightweight mesh
  structures}.
\newblock 3D Printing and Additive Manufacturing \textbf{2}(3), 117--122 (2015)

\bibitem{Huang2018automated}
Huang, Y., Garrett, C.R., Mueller, C.T.: {Automated sequence and motion
  planning for robotic spatial extrusion of 3D trusses}.
\newblock Construction Robotics \textbf{2}(1), 15--39 (2018).
\newblock \doi{10.1007/s41693-018-0012-z}

\bibitem{huang2016framefab}
Huang, Y., Zhang, J., Hu, X., Song, G., Liu, Z., Yu, L., Liu, L.: {Framefab:
  Robotic fabrication of frame shapes}.
\newblock ACM Transactions on Graphics (TOG) \textbf{35}(6), 224 (2016)

\bibitem{kingston2019exploring}
Kingston, Z., Moll, M., Kavraki, L.E.: {Exploring implicit spaces for
  constrained sampling-based planning}.
\newblock International Journal of Robotics Research \textbf{38}(10-11),
  1151--1178 (2019).
\newblock \doi{10.1177/0278364919868530}

\bibitem{KuffnerLaValle}
Kuffner~Jr., J.J., LaValle, S.M.: {{\{}RRT-Connect{\}}: An efficient approach
  to single-query path planning}.
\newblock In: IEEE International Conference on Robotics and Automation (ICRA)
  (2000)

\bibitem{lagriffoulbenchmarks}
Lagriffoul, F.: {On Benchmarks for Combined Task and Motion Planning}.
\newblock In: Robotics: Science and Systems (RSS) 2016 Workshop on Task and
  Motion Planning (2016)

\bibitem{lavalle1998rapidly}
LaValle, S.M.: {Rapidly-exploring random trees: A new tool for path planning}
  (1998)

\bibitem{McGuire_Gallagher_Ziemian_1999}
McGuire, W., Gallagher, R.H., Ziemian, R.D.: {Matrix Structural Analysis}.
\newblock Wiley (1999)

\bibitem{parascho_cooperative_2019}
Parascho, S.: Cooperative {Robotic} {Assembly}: {Computational} {Design} and
  {Robotic} {Fabrication} of {Spatial} {Metal} {Structures}.
\newblock Doctoral {Thesis}, ETH Zurich (2019).
\newblock \doi{10.3929/ethz-b-000364322}.
\newblock Accepted: 2019-09-17T07:16:57Z

\bibitem{parascho_computational_2018}
Parascho, S., Kohlhammer, T., Coros, S., Gramazio, F., Kohler, M.:
  Computational {Design} of {Robotically} {Assembled} {Spatial} {Structures}:
  {A} sequence based method for the generation and evaluation of structures
  fabricated with cooperating robots.
\newblock In: {AAG} 2018: {Advances} in {Architectural} {Geometry} 2018, pp.
  112--139. Klein Publishing (2018).
\newblock Accepted: 2018-10-29T05:56:52Z

\bibitem{schwartz2012hal}
Schwartz, T.: {HAL: Extension of a visual programming language to support
  teaching and research on robotics applied to construction}.
\newblock In: Robotic Fabrication in Architecture, Art and Design 2012, pp.
  92--101. Springer (2012)

\bibitem{soler2017generalized}
Soler, V., Retsin, G., Jimenez~Garcia, M.: {A Generalized Approach to
  Non-Layered Fused Filament Fabrication}.
\newblock Proceedings of the 36st Annual Conference of the Association for
  Computer Aided Design in Architecture (ACADIA) pp. 562--571 (2017)

\bibitem{sondergaard2016topology}
S{\o}ndergaard, A., Amir, O., Eversmann, P., Pi{\v{s}}korec, L., Stan, F.,
  Gramazio, F., Kohler, M.: {Topology optimization and robotic fabrication of
  advanced timber space-frame structures}.
\newblock In: Robotic Fabrication in Architecture, Art and Design 2016, pp.
  190--203. Springer (2016)

\bibitem{srivastava2014combined}
Srivastava, S., Fang, E., Riano, L., Chitnis, R., Russell, S., Abbeel, P.:
  {Combined Task and Motion Planning Through an Extensible Planner-Independent
  Interface Layer}.
\newblock In: IEEE International Conference on Robotics and Automation (ICRA)
  (2014)

\bibitem{stilman2010global}
Stilman, M.: {Global manipulation planning in robot joint space with task
  constraints}.
\newblock IEEE Transactions on Robotics \textbf{26}(3), 576--584 (2010)

\bibitem{sucan2013moveit}
Sucan, I.A., Chitta, S.: {Moveit!} (2018).
\newblock \urlprefix\url{http://moveit.ros.org}

\bibitem{tam2018}
Tam, K.M., Marshall, D.J.M., Gu, M., Kim, J., Huang, Y., Lavallee, J.A.,
  Mueller, C.T.: {Fabrication-aware structural optimisation of lattice
  additive-manufactured with robot-arm}.
\newblock International Journal of Rapid Manufacturing \textbf{7}(2-3) (2018)

\bibitem{thoma2018robotic}
Thoma, A., Adel, A., Helmreich, M., Wehrle, T., Gramazio, F., Kohler, M.:
  Robotic fabrication of bespoke timber frame modules.
\newblock In: Robotic Fabrication in Architecture, Art and Design, pp.
  447--458. Springer (2018)

\bibitem{toussaint2015logic}
Toussaint, M.: {Logic-geometric programming: an optimization-based approach to
  combined task and motion planning}.
\newblock In: IJCAI International Joint Conference on Artificial Intelligence,
  pp. 1930--1936. AAAI Press (2015)

\bibitem{willmann2016robotic}
Willmann, J., Knauss, M., Bonwetsch, T., Apolinarska, A.A., Gramazio, F.,
  Kohler, M.: Robotic timber construction—expanding additive fabrication to
  new dimensions.
\newblock Automation in construction \textbf{61}, 16--23 (2016)

\bibitem{wu2016printing}
Wu, R., Peng, H., Guimbreti{\`{e}}re, F., Marschner, S.: {Printing arbitrary
  meshes with a 5DOF wireframe printer}.
\newblock ACM Transactions on Graphics (TOG) \textbf{35}(4), 101 (2016)

\bibitem{yao2007path}
Yao, Z., Gupta, K.: {Path planning with general end-effector constraints}.
\newblock Robotics and Autonomous Systems \textbf{55}(4), 316--327 (2007)

\bibitem{yu2016acadia}
Yu, L., Huang, Y., Liu, Z., Xiao, S., Liu, L., Song, G., Wang, Y.: {Highly
  Informed Robotic 3D Printed Polygon Mesh: A Novel Strategy of 3D Spatial
  Printing}.
\newblock In: Proceedings of the 36st Annual Conference of the Association for
  Computer Aided Design in Architecture (ACADIA), pp. 298--307 (2016)

\bibitem{yuan2016robotic}
Yuan, P.F., Meng, H., Yu, L., Zhang, L.: {Robotic Multi-dimensional Printing
  Based on Structural Performance}.
\newblock In: Robotic Fabrication in Architecture, Art and Design 2016, pp.
  92--105. Springer (2016)

\end{thebibliography}


%
%

\end{document}